\documentclass{article}
\usepackage{spconf,amsmath,graphicx,hyperref}
\usepackage{amssymb,bm,amsthm}
\usepackage{algorithm,algorithmic}
\usepackage{xcolor}

\title{DPMM-CFL: Clustered Federated Learning via Dirichlet Process Mixture Model nonparametric clustering}

\name{
Mariona Jaramillo-Civill,
Peng Wu,
Pau Closas}

\address{Dept. of Electrical \& Computer Engineering, Northeastern University, Boston, MA, USA}

\begin{document}
\ninept
\maketitle
\begin{abstract}
Clustered Federated Learning (CFL) improves performance under non-IID client heterogeneity by clustering clients and training one model per cluster, thereby balancing between a global model and fully personalized models. However, most CFL methods require the number of clusters $K$ to be fixed a priori, which is impractical when the latent structure is unknown. 
We propose DPMM-CFL, a CFL algorithm that places a Dirichlet Process (DP) prior over the distribution of cluster parameters. 
This enables nonparametric Bayesian inference to jointly infer both the number of clusters and client assignments, while optimizing per-cluster federated objectives. 
This results in a method where, at each round, federated updates and cluster inferences are coupled, as presented in this paper.
The algorithm is validated on benchmark datasets under Dirichlet and class-split non-IID partitions.

\end{abstract}


\section{Introduction}
\label{sec:intro}
Federated learning (FL) enables collaborative model training across distributed clients under the coordination of a central server, while preserving data privacy and improving communication efficiency~\cite{mcmahan_communication-efficient_2017}. However, client data distributions are often heterogeneous (non-IID), making simple aggregation strategies with a single global model (e.g., FedAvg~\cite{mcmahan_communication-efficient_2017}) suboptimal. To address this, recent research has focused on personalized FL (PFL)~\cite{tan_towards_2023,kairouz_advances_2021,ma_convergence_2022}. Client-wise PFL assumes that each client’s data is unique and thus requires a dedicated local model (e.g., FedProx~\cite{li_federated_2018}, FedOpt~\cite{reddi_adaptive_2020}). To find a balance between fully local and global models, clustered FL (CFL) partitions clients into groups with similar data distributions and learns one model per cluster. Such CFL approaches have proven effective in diverse domains, ranging from handling heterogeneous data across edge devices~\cite{zhang_lcfed_2025} to mitigating inter-hospital variability in smart healthcare~\cite{ye_personalized_2025}.

Current CFL implementations follow different strategies to handle client heterogeneity. Federated Stochastic EM (FeSEM)~\cite{long_multi-center_2023} performs a K-means based CFL by measuring distances in the space of model parameters. In IFCA, each client selects the cluster whose model yields the lowest loss on its local data~\cite{ghosh_efficient_2022}. However, a key limitation of these approaches is that the number of clusters $K$ must be specified in advance. This requirement is unrealistic in practical FL scenarios, where the number of latent client groups is unknown and may evolve dynamically over rounds. 
Alternative frameworks such as hierarchical clustering~\cite{briggs_federated_2020} and cosine-similarity-based recursive partitioning~\cite{sattler_clustered_2019} have also been proposed, where clusters are iteratively merged or split based on pairwise similarity, thereby indirectly adapting $K$ during training. However, these methods depend on threshold hyperparameters for merging or splitting.  

In this work, we remove the need to fix $K$ by incorporating a Bayesian nonparametric framework into CFL. Specifically, we place a Dirichlet Process prior over cluster parameters, so the effective number of clusters and client assignments are inferred directly from data. This yields a threshold-free clustering mechanism driven by posterior inference, allowing clusters to split and merge adaptively during training. Our contribution is thus a CFL formulation that couples Bayesian inference over cluster structure with standard federated optimization to learn one model per inferred cluster. Section \ref{sec:method} presents the proposed CFL algorithm and how it is tightly coupled with the Bayesian clustering solution. Section \ref{sec:nonparam} provides theoretical and implementation details of the clustering method and Section \ref{sec:results} discusses experimental results validating the proposed method. The article concludes with final remarks in Section \ref{sec:conclusions}.

\section{Clustered federated learning with Bayesian clustering}\label{sec:method}

We aim to solve a clustered federated learning (CFL) problem in which the number of clusters is unknown and must be inferred jointly with the optimization of per-cluster models. This leads to two coupled problems, namely: (i) federated learning of global cluster models based on local client updates given a particular clustering assignment, whereby each client is aggregated with the updates of other clients associated to its same cluster; and (ii) clustering of clients based on their local weight updates, which we tackle as a nonparametric Bayesian inference problem where both assignments and number of clusters are unknown. The rest of the section provides more details about these problems as well as how they are coupled within the proposed method, referred to as DPMM-CFL.

\medskip
\noindent\textbf{(i) Federated optimization given cluster assignments.}  
We consider $M$ clients. Each client $i \in \{1,\dots,M\}$ holds a local dataset $\mathcal{D}_i$ with $n_i$ samples, and the total number of samples across all clients is $N = \sum_{i=1}^M n_i$.  
We define the per-client empirical loss $f_i(\cdot) := \ell(\mathcal{D}_i, \cdot)$ as the empirical loss of the model on dataset $\mathcal{D}_i$.  
Thus, the collection of client losses is denoted by $\{f_i(\cdot)\}_{i=1}^M$.
Each client $i$ is assigned to a cluster index $c_i \in \{1,\dots,K\}$, $K$ represents the number of clusters, which is unknown. The collection of cluster assignments forms the vector $\mathbf{c} = (c_1,\dots,c_M)$.
Equivalently, clustering assignments can be encoded in the binary matrix \(\mathbf{C} \in \{0,1\}^{M\times K}\) with \(r_{ik} = \mathbb{I}\{c_i = k\}\) in its $(i,k)$-th element.  
Each cluster \(k\in \{1,\dots,K\}\) maintains a model parameterized by \(\bm{\Omega}_k \in \bm{\Theta}\), and the collection of all cluster models is denoted by \(\bm{\Omega} = (\bm{\Omega}_1,\dots,\bm{\Omega}_K)\). 
In the clustered setting, the empirical risk minimization objective is given by \cite{ma_convergence_2022}:
\begin{equation}
\label{eq:cl_obj}
F(\bm{\Omega}; \mathbf{C}) \;=\; \sum_{k=1}^K \sum_{i=1}^M \frac{n_i}{N_k} r_{ik}\, f_i(\bm{\Omega}_k)\;,
\quad N_k = \sum_{i=1}^M r_{ik} n_{i},
\end{equation}
where $f_i(\bm{\Omega}_k)$ is the local loss of client $i$ evaluated with the parameters of cluster $k$.

We denote by \(\bm{\omega}_i\) the local parameter vector of client \(i\), 
and by \(\{\bm{\omega}_i\}_{i=1}^M\) the collection of all client parameters. Each client \(i\) initializes its local parameters with the model parameters of the cluster it has been associated with,  
\(\bm{\omega}_i^{0} = \bm{\Omega}_{c_i}\),  
and obtains updated parameters through local optimization:
\begin{equation}
 \bm{\omega}_i \;=\; \mathrm{LocalUpdate}(\bm{\omega}_i^{0};f_i,\mathcal{D}_i) \;.   
\end{equation}

Federated aggregation within cluster \(k\) then yields the usual federated averaging solution where parameters of clients in the same cluster are averaged together to produce the global model parameters for that cluster:
\begin{equation}
\label{eq:agg}
\bm{\Omega}_k \;=\; \sum_{i:\,c_i=k} \frac{n_i}{N_k} \bm{\omega}_i \;.
\end{equation}
Notice that the above FL local model update and aggregation is conditioned to a particular clustering solution. In practice, FL is implemented as iterative rounds where local and global updates are iteratively performed, in our case requiring an intermediate step of clustering clients based on their local weights. Our clustering approach is explained next.

\medskip
\noindent\textbf{(ii) Nonparametric Bayesian inference of cluster assignments with unknown $K$.}  
Given the set of updated client parameters \(\{\bm{\omega}_i\}_{i=1}^M\), we infer the latent partition structure \(\mathbf c\) that essentially clusters clients. We propose to use a DPMM to enable nonparametric Bayesian inference of both assignments $\mathbf{c}$ and number of clusters $K$.   
We consider a Dirichlet Process prior $G \sim \mathrm{DP}(\alpha,G_0)$ 
over cluster parameters, where $\alpha > 0$ is a concentration parameter 
controlling how often new clusters are created and $G_0$ is the base distribution. Formally, the DP supports infinitely many clusters, but in practice, with $M$ clients, the number of nonempty clusters is bounded by $M$.
The likelihood for the $i$-th client is defined as
\begin{equation}
\bm{\omega}_i \mid c_i,\bm{\mu}_{c_i} \sim p(\bm{\omega}_i \mid \bm{\mu}_{c_i}),
\quad \bm{\mu}_k \sim G_0.
\end{equation}
Integrating out both \(G\) and $\bm{\mu}_k$, we obtain the marginalized posterior over cluster assignments:
\begin{align}
p(\mathbf c \mid \bm{\omega}_{1:M}, \alpha, G_0) 
& \propto p(\mathbf c \mid \alpha)\, \prod_{k=1}^{K} p(\bm{W}_k \mid G_0)
\end{align}
where \(\bm{W}_k=\{\bm{\omega}_i:c_i=k\}\) and \(p(\bm{W}_k\mid G_0)\) denotes the marginal 
likelihood of cluster \(k\).
Since no closed form exists to compute point estimates of the cluster assignments when the number of clusters is unknown, we resort to MCMC to sample from the marginal posterior of $\mathbf{c}$.
The full derivation of this posterior and its sampling-based inference procedure, is introduced in Sec.~\ref{sec:nonparam}.

\medskip
\noindent\textbf{Algorithm description.}
The two problems are tightly coupled and give rise to an iterative procedure over communication rounds in the usual federated learning approaches. 
Algorithm~\ref{alg:dpmm_cfl} summarizes the relevant operations involved in the method.
Unlike static clustering, client representations (that is, the local parameters $\{\bm{\omega}_i\}$)
are continually updated each round, initialized from the current cluster models $\bm{\Omega}$ and refined through local Stochastic Gradient Descent (SGD) steps. The main steps in this iterative algorithm are:
\begin{itemize}
    \item \emph{Initial broadcast:} at the beginning, all clients belong to a single cluster, $K=1$. The server initializes this cluster model $\bm{\Omega}_1^0$ randomly and broadcasts it to every client.
  \item \emph{Local update:} each client initializes $\bm{\omega}_i^{t,0} \!\leftarrow\! \bm{\Omega}_{c_i^{t-1}}^{\,t-1}$ and runs $Q$ steps of local SGD on its local dataset $\mathcal{D}_i$ to obtain $\bm{\omega}_i^t$.
  \item \emph{DPMM clustering:} using the updated $\{\bm{\omega}_i^t\}$, the server samples new assignments
  $\mathbf c^t \sim p(\mathbf c \mid \bm{\omega}_{1:M}^t,\alpha,G_0)$ via split--merge MCMC sampling of the joint posterior of cluster assignments, which
  implies a (possibly) new number of clusters $K_t$ at every round.
  \item \emph{Aggregation and distribution:} for each cluster $k \le K_t$, the server aggregates the model updates from corresponding clients, 
  $\bm{\Omega}_k^{t} = \sum_{i:\,c_i^{\,t}=k} \frac{n_i}{N_k^{t}} \bm{\omega}_i^{t}$, where $N_k^{t}=\sum_{i:\,c_i^{\,t}=k} n_i$, and redistributes $\bm{\Omega}_k^{t}$ among those clients.
\end{itemize}

Notice that cluster inference uses the most recent client updates $\{\bm{\omega}_i^t\}_{i=1}^M$, while federated
optimization conditions on the sampled clustering assignment $\mathbf c^t$ (and $K_t$) through the objective
$F(\bm{\Omega};\mathbf{C})$ and the aggregation rule. This coupled problem is solved iteratively until convergence.

\begin{algorithm}[t]
\caption{Dirichlet Process Mixture Model -- Clustered FL (DPMM\text{-}CFL)}
\label{alg:dpmm_cfl}
\begin{algorithmic}[1]
\STATE \textbf{Input:} client datasets $\{\mathcal{D}_i\}_{i=1}^M$, loss functions $\{f_i\}_{i=1}^M$, DP hyperparameters $(\alpha, G_0)$, local steps $Q$, rounds $T$
\STATE \textbf{Initialize:} set $K_0 = 1$, random cluster model $\bm{\Omega}_1^0$ and assignments $\{c_i^0=1\}_{i=1}^M$
\FOR{$t=1,\dots,T$}
  \STATE \textbf{Local update step (all clients $i=1,\dots,M$):}
  \STATE \hspace{1.0em} Initialize $\bm{\omega}_i^{t,0} \!\leftarrow\! \bm{\Omega}_{c_i^{t-1}}^{\,t-1}$
  \STATE \hspace{1.0em} Run $Q$ steps of SGD on local data $\mathcal{D}_i$ to reduce the clustered objective $F(\bm{\Omega};C)$ in \eqref{eq:cl_obj} (now $K = K_{t-1}$)
  \[
      \bm{\omega}_i^t \;=\; \mathrm{LocalUpdate}_Q(\bm{\omega}_i^{t,0}; f_i, \mathcal{D}_i).
  \]

  \STATE \textbf{DPMM clustering step:}
  \STATE \hspace{1.0em} Sample new assignments via split--merge MCMC:
  \[
      \mathbf c^t \sim p(\mathbf c \mid \bm{\omega}_{1:M}^t, \alpha, G_0).
  \]
\vspace{-1.4em}
  \STATE \hspace{1.0em} Set $K_t \leftarrow$ number of distinct clusters in $\mathbf c^t$

  \STATE \textbf{Aggregation and distribution step:}
  \FOR{each $k=1,\dots,K_t$}
    \STATE \hspace{1.0em} Set $N_k^{t} \leftarrow \sum_{i:\,c_i^{\,t}=k} n_i$
    \STATE \hspace{1.0em} Update cluster model (\eqref{eq:agg}): $\bm{\Omega}_k^{t} \leftarrow \sum_{i:\,c_i^{\,t}=k} \frac{n_i}{N_k^{t}} \bm{\omega}_i^{t}$
    \STATE \hspace{1.0em} Send $\bm{\Omega}_k^t$ to all clients with $c_i^t = k$
\vspace{-0.3em}
  \ENDFOR
\ENDFOR

\STATE \textbf{Output:} cluster assignments $\mathbf c^{t=T}$, final number of clusters $K_{t=T}$, cluster models $\bm{\Omega}^{T}=(\bm{\Omega}^{t=T}_1,\dots,\bm{\Omega}^{t=T}_{K_T})$

\end{algorithmic}
\end{algorithm}

\section{Nonparametric Clustering of Federated Learning clients with DPMM}
\label{sec:nonparam}
Since the number of clusters is not known a priori, we treat client clustering as a Bayesian nonparametric inference problem and model it with a Dirichlet Process Mixture Model (DPMM).

A Dirichlet process (DP), \(G \sim \mathrm{DP}(\alpha,G_0)\), is a distribution over probability measures~\cite{ferguson_bayesian_1973}, 
characterized by a concentration parameter \(\alpha>0\), controlling the creation of new clusters, and a base distribution \(G_0\), 
acting as the prior over cluster-specific parameters~\cite{theodoridis_machine_2020}. Its defining property is that for any finite partition 
\(T_1,\dots,T_K\) of the parameter space \(\bm{\Theta}\),
\begin{equation}
(G(T_1),\dots,G(T_K)) \sim \mathrm{Dir}\!\left(\alpha G_0(T_1),\dots,\alpha G_0(T_K)\right) \;.
\end{equation}

A DPMM~\cite{antoniak_mixtures_1974} is defined hierarchically:
\begin{equation}
G \sim \mathrm{DP}(\alpha, G_0),
\end{equation}
\begin{equation}
\bm{\theta}_i \mid G \sim G,
\end{equation}
\begin{equation}
\bm{\omega}_i \mid \bm{\theta}_i \sim p(\bm{\omega}_i \mid \bm{\theta}_i), \quad i=1,\dots,M.
\end{equation}
Since draws from \(G\) are almost surely discrete~\cite{ferguson_prior_1974}, multiple \(\bm{\theta}_i\) coincide, 
inducing clusters without fixing their number in advance. Here \(G_0\) specifies where cluster parameters lie, 
while \(\alpha\) balances reuse of existing clusters versus creation of new ones.

Integrating out \(G\) yields the Pólya urn model~\cite{blackwell_ferguson_1973}, 
where new clusters arise with probability proportional to \(\alpha\). 
For cluster assignments we use the equivalent Chinese Restaurant Process (CRP):  
\begin{equation}
P(c_i = k \mid c_{1:i-1}, \alpha) =
\begin{cases}
\dfrac{m_k^{(i-1)}}{\,i-1+\alpha\,}, & \text{if } k \le K_{i-1}, \\[8pt]
\dfrac{\alpha}{\,i-1+\alpha\,}, & \text{otherwise },
\end{cases}
\end{equation}
where \(K_{i-1}\) is the number of occupied clusters after the first \(i-1\) observations, 
and \(m_k^{(i-1)}\) is the number of those \(i-1\) observations already assigned to cluster \(k\).



Multiplying the sequential conditionals over all \(M\) observations and using exchangeability yields the CRP joint prior, which does not depend on the order in which the observations are considered~\cite{antoniak_mixtures_1974}:
\begin{eqnarray}
\label{eq:priorcrp}
p(\mathbf c \mid \alpha) &
\;\overset{\text{\cite{jain_split-merge_2004}}}{=}\; &
\frac{\alpha^{K}\prod_{k=1}^{K} (m_k-1)!}{\prod_{i=1}^{M} (\alpha+i-1)} \nonumber \\
& \;\overset{\text{\cite{antoniak_mixtures_1974}}}{=}\; &
\frac{\Gamma(\alpha)}{\Gamma(\alpha+M)}\,\alpha^{K}\prod_{k=1}^{K} \Gamma(m_k),
\end{eqnarray}
where \(\mathbf c=(c_1,\dots,c_M)\) is the assignment vector, \(K\) the number of occupied clusters, 
and \(m_k\) their sizes.

For this work, we adopt a Gaussian base distribution 
\(G_0 = \mathcal N(\bm{\mu}_0, \bm{\Sigma}_0)\), which serves as a prior for the 
cluster parameters in our likelihood model. The likelihood is also Gaussian, 
\(\bm{\omega}_i \mid \bm{\mu} \sim \mathcal N(\bm{\mu}, \bm{\Sigma})\), yielding an infinite Gaussian 
mixture model~\cite{rasmussen_infinite_2000} with Normal–Normal conjugacy and hyperparameter $\bm{\Sigma}$. Our goal is then to infer the posterior distribution over cluster assignments:
\begin{align}
\label{eq:posterior_c}
p(\mathbf c \mid \bm{\omega}_{1:M}, \alpha, G_0) 
& \propto p(\mathbf c \mid \alpha)\, p(\bm{\omega}_{1:M} \mid \bm{c},G_0) \nonumber \\
&= p(\mathbf c \mid \alpha)\, \prod_{k=1}^{K} p(\bm{W}_k \mid G_0)
\end{align}
where \(\bm{W}_k=\{\bm{\omega}_i:c_i=k\}\) and \(p(\bm{W}_k\mid G_0)\) denotes the marginal 
likelihood of cluster \(k\). This marginal can be expressed as an 
integral over the cluster parameters,
\begin{equation}
\label{eq:likelihoodmarg}
p(\bm{W}_k \mid G_0) = \int \Bigg[\prod_{\bm{\omega}\in \bm{W}_k} \mathcal N(\bm{\omega}\mid \bm{\mu},\bm{\Sigma})\Bigg]\,
\mathcal N(\bm{\mu}\mid \bm{\mu}_0,\bm{\Sigma}_0)\,d\bm{\mu},
\end{equation}
which has a closed form under Normal–Normal conjugacy. While the exact expression 
is available in the literature (see~\cite{demichelis_hierarchical_2006}), 
we omit it here since it is only required for implementation details.  

Our goal is to infer the cluster assignments \(\mathbf c\), i.e., to sample from the posterior 
\(p(\mathbf c \mid \bm{\omega}_{1:M}, \alpha, G_0)\). Since no closed form exists when the number of clusters 
is unknown, we resort to MCMC. One option is Gibbs sampling, which updates each 
assignment \(c_i\) in turn, but its incremental nature often traps the chain in local modes. To address this, we use split–merge MCMC with 
\textit{Restricted Gibbs Sampling Proposals from a Random 
Launch State} (Sec.~3.3.2 in~\cite{jain_split-merge_2004}), updating groups of observations simultaneously. 
A proposed split or merge is then accepted with Metropolis–Hastings probability~\cite{murphy_machine_2012}:
\begin{equation}
a(\mathbf c^\star, \mathbf c) 
= \min\!\left\{1,\,
\frac{q(\mathbf c \mid \mathbf c^\star)}{q(\mathbf c^\star \mid \mathbf c)} \cdot
\frac{p(\mathbf c^\star \mid \bm{\omega}_{1:M}, \alpha, G_0)}
     {p(\mathbf c \mid \bm{\omega}_{1:M}, \alpha, G_0)}
\right\}
\end{equation}
where \(q(\cdot\mid\cdot)\) is the proposal distribution in~\cite{jain_split-merge_2004}, 
and the posterior \(p(\mathbf c \mid \bm{\omega}_{1:M}, \alpha, G_0)\) is given in~\eqref{eq:posterior_c}--\eqref{eq:likelihoodmarg}.

\vspace*{-.25cm}
\section{Evaluation} \label{sec:results}

\begin{figure*}[!htbp]
\centering
  \includegraphics[width=\textwidth]{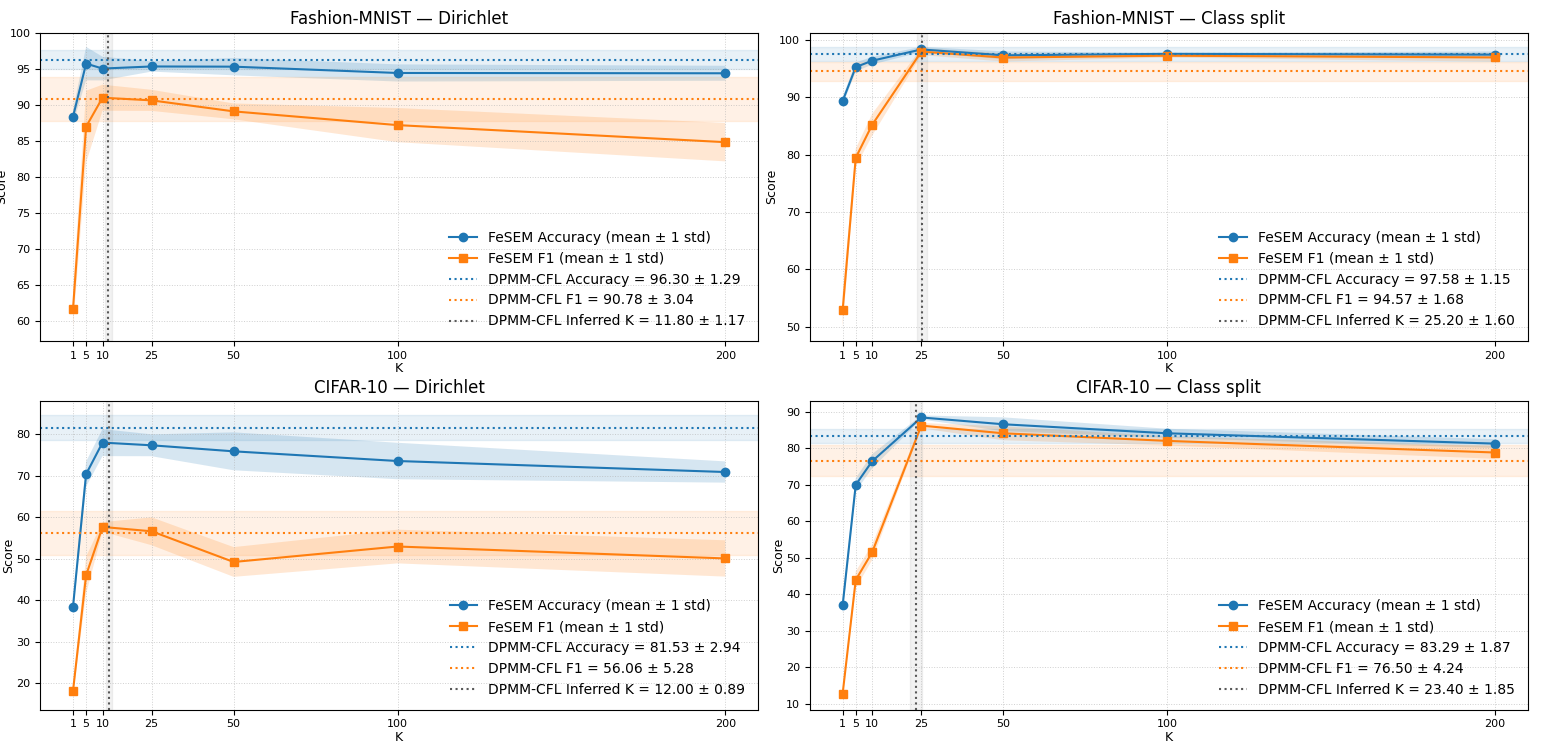}
  \caption{Performance of FeSEM (fixed $K$) versus DPMM-CFL (unknown $K$) on Fashion-MNIST and CIFAR-10 
  under Dirichlet and class-split partitions. Solid curves with shaded regions show mean $\pm$1 standard 
  deviation across seeds. Dotted horizontal lines indicate mean performance for the unknown $K$ case, and 
  dotted vertical lines indicate the mean inferred number of clusters (with variability across seeds).}
  \label{fig:results}
\end{figure*}

\vspace*{-.25cm}
\subsection{Datasets and Non-IID generation}
We evaluate the proposed methodology on two standard benchmarks:  
1) Fashion-MNIST~\cite{xiao_fashion-mnist_2017}: 70,000 grayscale images ($28 \times 28$ pixels) evenly split across 10 classes, with 60,000 training and 10,000 test samples;
2) CIFAR-10~\cite{krizhevsky_learning_2012}: 60,000 color images ($32 \times 32$ pixels) in 10 classes, with 50,000 training and 10,000 test samples. CIFAR-10 exhibits much stronger heterogeneity compared to MNIST-family datasets.  

To simulate cluster-wise non-IID conditions, we apply two preprocessing methods, following~\cite{ma_convergence_2022, wu_bayesian_2024}:  

\emph{Dirichlet partitioning}: This method uses the Dirichlet distribution to control the degree of non-IIDness. The dataset is first split into $K=10$ clusters with concentration parameter $\alpha=0.1$, inducing strong inter-cluster variability. Each cluster is then further divided into $m/K$ clients with $\alpha=10$ to control intra-cluster heterogeneity. This setup yields a clear clustering structure, since the number of ground-truth clusters is fixed at $K=10$.    

\emph{Class-split partitioning}: Following FedAvg~\cite{mcmahan_communication-efficient_2017}, each cluster is assigned a random subset of 3 classes, with a relatively balanced number of samples per class, and each client receives data from 2 of these classes. Unlike the Dirichlet case, the underlying clustering structure is more complex for CIFAR-10, where the true number of latent clusters is unknown.  

\vspace*{-.25cm}
\subsection{System Settings}
We simulate $200$ clients, each using a CNN~\cite{lecun_deep_2015} where the final fully connected layer serves as the client representation for clustering~\cite{ma_convergence_2022}. Training uses stochastic gradient descent (SGD) with learning rate $0.005$ and momentum $0.9$, batch size $32$, and local steps $Q=10$. The objective function is cross-entropy loss. We run $T=200$ communication rounds.  

For the Bayesian nonparametric model, the DP concentration parameter is set to $\alpha=1.0$ (weakly informative), and the base measure $G_0$ is a spherical Gaussian $\mathcal N(\bm{\mu}_0,{\sigma}_0 \mathbf{I})$ with $\bm{\mu}_0=\bm{0}$ and ${\sigma}_0^2=1$. Results are reported in terms of micro-accuracy ($\%$) and macro-F1, both averaged across clients. Each experiment is repeated over five Monte Carlo runs with different random seeds. Mean values are reported from the last three rounds, and shaded regions in Fig.~\ref{fig:results} indicate one standard deviation.   

\vspace*{-.25cm}
\subsection{Results}
We compare our proposed DPMM-CFL against FeSEM~\cite{long_multi-center_2023}, a K-means-based CFL baseline with fixed $K$. For fairness, FeSEM is run with multiple choices of $K$, while DPMM-CFL performs Bayesian inference of $K$. We evaluate four scenarios, combining the two datasets with the two preprocessing methods, and report results in terms of converged accuracy and F1 score in Fig.~\ref{fig:results}.

On Fashion-MNIST and CIFAR-10 under Dirichlet partitioning, DPMM-CFL outperforms FeSEM for almost all fixed $K$ values. Importantly, DPMM-CFL infers $\sim 12$ clusters on average, close to the true $K=10$ used for generating the data.  

On Fashion-MNIST and CIFAR-10 under class-split partitioning, the ground-truth clustering is unknown. Here, the number of clusters inferred by DPMM-CFL aligns with the $K$ at which FeSEM reaches peak performance (around 25 clusters). Although DPMM-CFL slightly underperforms FeSEM at high $K$ (beyond 20), previous studies only evaluated FeSEM up to $K \leq 10$ due to the computational cost of sweeping over all $K$, thereby reporting worse performance~\cite{ma_convergence_2022}.
Thus, DPMM-CFL provides a principled approach by automatically inferring both the number of clusters and client assignments, while avoiding exhaustive sweeps over possible $K$.

\begin{figure}[t]
  \centering
  \includegraphics[width=\linewidth]{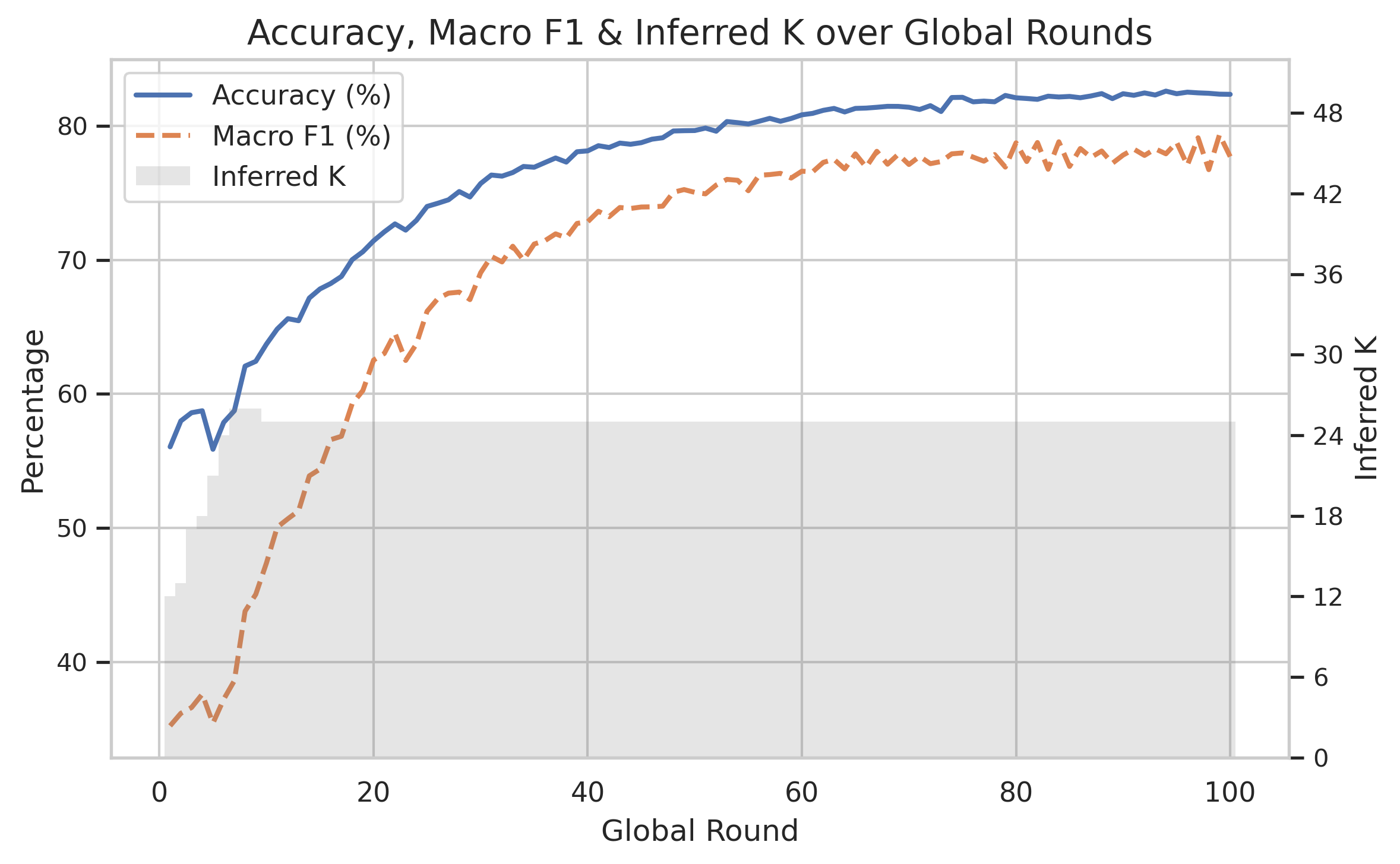}
  \caption{Empirical convergence of DPMM-CFL on CIFAR-10 with class-split (3 classes per cluster, 2 classes per client): accuracy and F1 score converge while the inferred $K_t$ stabilizes after early rounds.}
\vspace*{-.25cm}
  \label{fig:convergence}
\end{figure}

We analyze empirical convergence in terms of accuracy, F1 score, and the inferred number of clusters. For stable training, it is crucial that $K_t$ stabilizes early so that cluster assignments do not fluctuate across rounds. 
A representative Monte Carlo run is shown in Fig.~\ref{fig:convergence} to illustrate this behavior: the number of clusters initially grows and decreases through splitting and merging moves, showing that the sampling algorithm is functioning properly. After this transient phase, $K_t$ stabilizes and client assignments remain consistent, enabling the algorithm to learn per-cluster models until accuracy and F1 score converge.

\vspace*{-.25cm}
\section{Conclusions}
\label{sec:conclusions}

We introduced DPMM\text{-}CFL, a clustered federated learning framework that removes the need to fix the number of clusters a priori by leveraging a Dirichlet Process mixture model. The method jointly infers cluster assignments and the effective number of clusters from client updates while optimizing per-cluster objectives through local SGD and within-cluster aggregation. We conducted a thorough evaluation on Fashion\-MNIST and CIFAR-10 under both Dirichlet and class-split non-IID regimes, covering scenarios where the true number of clusters is known (Dirichlet) and unknown (class-split). In the former, the inferred number closely matches the ground truth, while in the latter, it aligns with the $K$ that yields peak accuracy and F1 score in FeSEM, the K-means-based CFL baseline. Finally, convergence analysis shows that the inferred number of clusters stabilizes early through splitting and merging, after which cluster assignments remain consistent and accuracy and F1 score empirically converge.

\vfill \pagebreak
\clearpage
\label{sec:refs}

\bibliographystyle{IEEEbib}
\bibliography{references}

\end{document}